\documentclass[letterpaper,journal]{IEEEtran}
\usepackage{amsmath,amsfonts}
\usepackage{algorithmic}
\usepackage{algorithm}
\usepackage{array}
\usepackage[caption=false,font=normalsize,labelfont=sf,textfont=sf]{subfig}
\usepackage{textcomp}
\usepackage{stfloats}
\usepackage{url}
\usepackage{verbatim}
\usepackage{graphicx}
\usepackage{cite}
\usepackage{booktabs}
\usepackage{multirow}
\usepackage{color}
\usepackage{hyperref}
\hypersetup{
    hidelinks,
    pdfborder={0 0 0},
    colorlinks=false,
    pdftitle={Track A*: Fast Visibility-Aware Trajectory Planning for Active Target Tracking},
    pdfauthor={Hanxuan Chen, Kangli Wang, Ji Pei},
    pdfkeywords={Active target tracking, spatio-temporal planning, A*, visibility-aware planning, UAV},
    pdfsubject={Trajectory planning for active target tracking}
}

\begin{document}

\title{Track A*: Fast Visibility-Aware Trajectory Planning for Active Target Tracking}

\author{Hanxuan Chen, Kangli Wang, and Ji Pei
\thanks{Hanxuan Chen, Kangli Wang, and Ji Pei are with Autel Robotics (e-mail: \{chenhanxuan, wangkangli, peiji\}@autelrobotics.com).}
\thanks{Corresponding author: Ji Pei.}
}

\maketitle

\begin{abstract}
Offline reference trajectories for active target tracking are needed both for building multi-modal tracking datasets and for benchmarking online tracking planners under repeatable conditions.
We present Track A* (TA*), an offline search-based trajectory planner that targets the visibility-aware target tracking objective on a discretized four-dimensional spatio-temporal grid $(x, y, z, t)$.
TA* combines a layered Directed Acyclic Graph (DAG) search with three engineering optimizations: cross-time obstacle distance caching against a Bounding Volume Hierarchy (BVH), per-layer beam pruning, and a configurable multi-ray visibility evaluator.
TA* employs a beam-pruned heuristic search on this discrete graph to efficiently find high-quality tracking trajectories. While it trades strict theoretical optimality for practical scalability, our empirical results demonstrate robust, near-baseline visibility performance at a fraction of the computational cost.
On a 1000-scenario stress test across eight CARLA Optimized maps, TA* converges on all scenarios and completes in 45~s using 32 workers; on a 248-scenario controlled comparison against an unoptimized priority-queue A* baseline (BinaryHeap implementation) under identical scenario inputs and a $5{\times}10^{6}$ expansion cap, TA* reduces mean planning time by 23.0$\times$ and worst-case planning time by 11.8$\times$, while raising convergence from 56.9\% to 100\%.
On the $n=141$ baseline-converged subset, TA* changes average visibility by only $-0.15$ percentage points (pp), with no scenario exceeding a 5~pp drop.
We position TA* as a practical offline reference planner under these specific conditions, with limitations and failure cases discussed for environments such as Town07 dense vegetation.
\end{abstract}


\section{Introduction}
\IEEEPARstart{A}{ctive} target tracking is a fundamental capability for autonomous robots, enabling applications ranging from cinematography and surveillance to embodied AI and autonomous driving \cite{bonatti2019towards, wang2021visibility}. The core challenge lies in generating a trajectory for the tracking robot (the "tracker") that maintains continuous visibility of a moving target while avoiding obstacles and ensuring dynamic feasibility.

While significant progress has been made in online, real-time local planning methods (e.g., Artificial Potential Fields \cite{khatib1986real}, Model Predictive Control \cite{falcone2007predictive}), these approaches are susceptible to local minima in complex, obstacle-dense environments. When a target moves behind a large building or into a narrow corridor, local planners may fail to find a path that restores visibility. Consequently, there is an academic and practical need for \textit{offline search-based trajectory planning}. Such planners serve two roles: (1) generating high-quality reference trajectories for multi-modal tracking datasets that can be used to train or evaluate learning-based embodied AI agents; and (2) providing a reproducible offline baseline against which online tracking algorithms can be evaluated.

Traditional global planners, such as A* \cite{hart1968formal} and RRT* \cite{karaman2011sampling}, are designed for static 2D or 3D spatial environments. Spatio-temporal search has been studied for autonomous driving and racing, where the goal is to navigate an ego vehicle around dynamic obstacles \cite{xin2021enable,rowold2022efficient,zhong2024spatio}. Target tracking, however, imposes a different objective: the planner must choose a tracker trajectory that remains time-synchronized with a moving target while preserving line-of-sight visibility. This requires searching a four-dimensional spatio-temporal space $(x, y, z, t)$ and redesigning the cost function around target position deviation, temporal smoothness, tracker kinematics, and continuous visibility.

Recent visibility-aware trackers, such as visPlanner \cite{wang2021visibility}, Eva-Tracker \cite{wang2023eva}, and SVPTO \cite{wang2023svpto}, have proposed innovative solutions for maintaining target visibility. However, these methods are predominantly local optimization, MPC, or learning-based planners. They are effective for online aerial tracking but do not provide an offline global search baseline that can systematically reason over all tracker states across time. A* has also been used in UAV target tracking, but primarily as a spatial obstacle-avoidance component rather than a 4D visibility-aware search over target motion \cite{cai2019path}. Recent surveys similarly identify optimization, sampling, and learning as the dominant paradigms, leaving search-based spatio-temporal tracking underexplored \cite{wu2025uav}.

In this paper, we propose Track A* (TA*), an offline search-based trajectory planner for visibility-aware target tracking under a configurable, viewpoint-agnostic objective. TA* addresses the gap between spatio-temporal A* methods for ego-navigation and visibility-aware local trackers by combining a 4D layered DAG search, a target-tracking-specific visibility objective, and practical offline optimizations. TA* employs a beam-pruned heuristic search on this discrete graph to efficiently find high-quality tracking trajectories; while it trades strict theoretical optimality for practical scalability, our empirical results demonstrate robust, near-baseline visibility performance at a fraction of the computational cost. Our large-scale evaluation focuses on aerial tracking at 22~m altitude, while the formulation itself is parameterized by altitude, camera distance, and viewpoint offsets and is therefore applicable to other tracker configurations not separately quantified in this work.

Our main contributions are:
\begin{enumerate}
    \item \textbf{4D Layered DAG Search for Tracking:} We lift A*-style search from static 2D/3D pathfinding to a four-dimensional $(x, y, z, t)$ layered DAG search that explicitly reasons about target motion and tracker timing, with a layer-local beam to bound work per layer.
    \item \textbf{Visibility-Aware Tracking Objective:} We formulate a task-specific cost function that combines tracking deviation, obstacle safety, line-of-sight visibility from a 5-ray sampler, and a vertical smoothness term.
    \item \textbf{Offline Reference Trajectory Generation:} We package the search and a beam-width default (\(B=2048\)) chosen for an empirical 5-pp worst-case visibility envelope on our 248-scenario subset, suitable for dataset construction and offline benchmarking.
    \item \textbf{Cross-Time Obstacle Distance Caching:} We share static obstacle distances across the time dimension via a Bounding Volume Hierarchy (BVH) and evaluate TA* on 1000 CARLA tracking scenarios, including a Town07 dense-vegetation failure case.
\end{enumerate}

\section{Related Work}

\subsection{Spatio-Temporal Search-Based Planning}
Search-based methods, particularly A* \cite{hart1968formal} and its variants such as Theta* \cite{nash2007theta} and D* Lite \cite{koenig2002d}, provide strong guarantees on discretized graphs. Recent work has extended graph search to spatio-temporal planning for autonomous vehicles. Xin et al. \cite{xin2021enable} construct a DAG-based spatio-temporal driving space for faster and smoother autonomous vehicle maneuvers, Rowold and Betz \cite{rowold2022efficient} use efficient spatiotemporal graph search for local racing trajectory planning, and Zhong et al. \cite{zhong2024spatio} combine search and optimization for autonomous driving. These methods demonstrate the value of time-indexed search, but their objective is ego-navigation and dynamic obstacle avoidance rather than maintaining visibility of a moving target.

\subsection{Visibility-Aware Target Tracking}
Visibility-aware trajectory planning is an active research area, especially for UAV tracking. visPlanner \cite{wang2021visibility} formulates aerial tracking as a visibility-aware trajectory optimization problem using ESDFs, while Eva-Tracker \cite{wang2023eva} removes the ESDF dependency through ray-casting and MPC. SVPTO \cite{wang2023svpto} further incorporates safe visibility-guided perception-aware optimization, and recent UAV-tracking surveys document a parallel line of learning-based occlusion-aware active trackers \cite{wu2025uav}. These approaches target online planning and perception-aware control, but they remain local or learning-based and therefore do not provide a search-based offline global baseline.

\subsection{A* for Target Tracking and the Remaining Gap}
A* has occasionally appeared in target tracking systems. For example, Cai et al. \cite{cai2019path} use an improved A* algorithm for UAV path planning during target tracking, but the search is spatial and is coupled with a separate tracking controller; it does not search over time or optimize a visibility-aware tracking objective. Recent surveys of UAV target tracking identify optimization-based, sampling-based, and learning-based planners as the dominant families, without identifying a 4D A*-based planner designed for active target tracking \cite{wu2025uav}. We are not aware of prior work that jointly addresses search-based 4D spatio-temporal planning, explicit line-of-sight visibility optimization, target-following objectives, and offline reference trajectory generation; Table~\ref{tab:related_axes} summarizes this gap on five comparison axes. Entries in the table reflect the dominant claim of each cited paper rather than its full feature set, and ``--'' marks an axis not addressed by that paper. TA* is designed to address this gap under the discrete-graph proximal-optimal objective defined in \S\ref{sec:method} and to support dataset generation and benchmarking under a configurable viewpoint specification. Table~\ref{tab:related_axes} is a coarse taxonomy for situating TA* and is not intended as a complete feature audit of the cited works.

\begin{table*}[t]
\caption{Representative tracking and spatio-temporal planners on five axes relevant to TA*.}
\label{tab:related_axes}
\centering
\small
\begin{tabular}{lccccc}
\toprule
\textbf{Method} & \textbf{Search-based} & \textbf{4D $(x,y,z,t)$} & \textbf{Visibility objective} & \textbf{Target tracking} & \textbf{Offline / global} \\
\midrule
visPlanner \cite{wang2021visibility}        & --        & --        & yes (ESDF)        & yes & local \\
Eva-Tracker \cite{wang2023eva}              & --        & --        & yes (ray-cast)    & yes & local \\
SVPTO \cite{wang2023svpto}                  & --        & --        & yes (perception)  & yes & local \\
Cai et al.\ \cite{cai2019path}              & yes (A*)  & no (3D)   & no                & yes & spatial \\
Xin et al.\ \cite{xin2021enable}            & yes (DAG) & yes       & no                & no (ego-nav) & global \\
Rowold and Betz \cite{rowold2022efficient}  & yes       & yes       & no                & no (racing)  & global \\
Zhong et al.\ \cite{zhong2024spatio}        & yes       & yes       & no                & no (driving) & global \\
\midrule
TA* (ours)                                  & yes       & yes       & yes (5-ray LoS)   & yes & offline \\
\bottomrule
\end{tabular}
\end{table*}

\begin{figure}[t]
\centering
\includegraphics[width=\columnwidth]{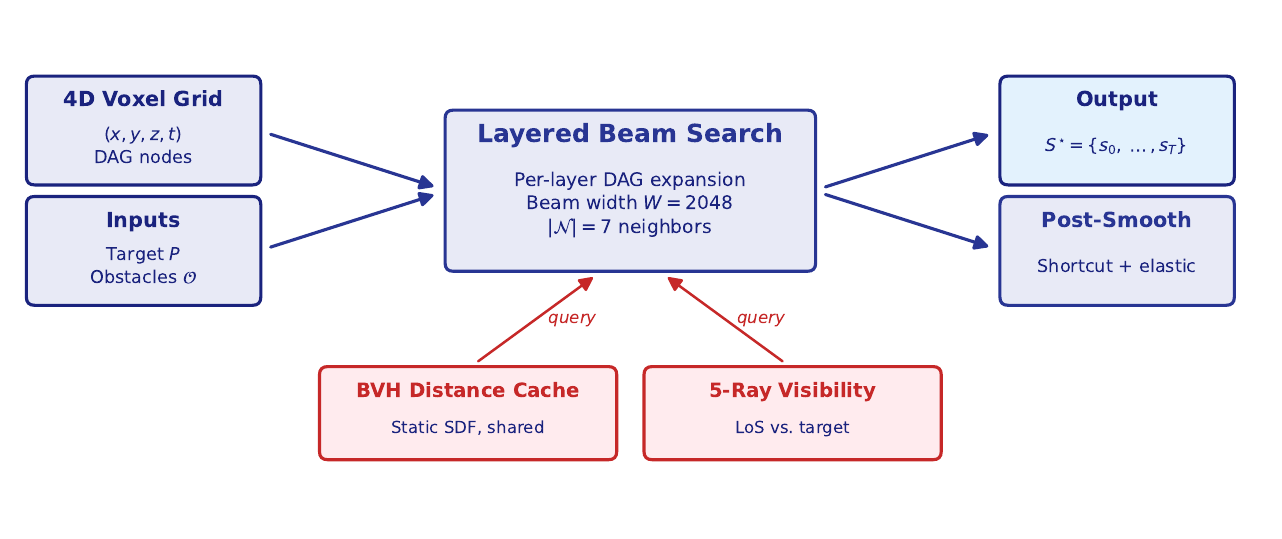}
\caption{Overview of the TA* algorithm pipeline (\S\ref{sec:method}); the
discretisation, neighbour set, beam width, distance cache, and 5-ray
visibility evaluator are described in the cited subsections.}
\label{fig:pipeline}
\end{figure}

\section{The Track A* (TA*) Algorithm}
\label{sec:method}

\subsection{Problem Formulation}
\label{subsec:formulation}

\paragraph{Inputs} The planner receives: (i) a discrete target trajectory
$P = \{p_t\}_{t=0}^{T} \subset \mathbb{R}^3$ sampled at uniform time step $\Delta t$,
(ii) a static obstacle set $\mathcal{O}$ represented as oriented bounding boxes and
queried through a Bounding Volume Hierarchy (BVH), and (iii) an initial tracker
position $s_0 \in \mathbb{R}^3$. We use $\Delta t = 0.5$~s in all experiments
(constant-speed pedestrian model at $1.4$~m/s). The horizon $T$ is set by the
target trajectory length and corresponds to a 150--400~m walking path.

\paragraph{State and decision space} A tracker state at layer $t$ is
$s_t = (x_t, y_t, z_t)$ in world coordinates. We do not include velocity in
the state: any feasible motion uses voxel-aligned positions, so velocity is
the derived quantity $v_t = (s_t - s_{t-1}) / \Delta t$. The continuous space is
discretized by a uniform voxel grid with horizontal resolution
$\Delta_{xy} = 4$~m and vertical resolution $\Delta_{z} = 4$~m. Search is
restricted to a corridor of width $\pm 45$~m around the target path, plus
buffer space around obstacles whose footprint enters this corridor.

\paragraph{Feasibility constraints} A state $s_t$ is feasible if
\begin{align}
\label{eq:feasible}
&z_{\min} \le z_t \le z_{\max}, \\
&d_{\min}^{\mathrm{cam}} \le \|s_t - p_t\| \le d_{\max}^{\mathrm{cam}}, \\
&\mathrm{dist}(s_t, \mathcal{O}) \ge d_{\mathrm{safe}},
\end{align}
and an edge $(s_{t-1}, s_t)$ is admissible when
$\|s_t - s_{t-1}\| \le v_{\max}\,\Delta t$. We use
$d_{\min}^{\mathrm{cam}} = 3$~m, $d_{\max}^{\mathrm{cam}} = 50$~m,
$d_{\mathrm{safe}} = 1.5$~m, and $v_{\max} = 10$~m/s. Edge collision is
checked only at the endpoints; obstacle bounding boxes are inflated by
$d_{\mathrm{safe}}$ prior to BVH queries, which together with the voxel
resolution provides an implicit swept-volume margin in our experimental
setting. This endpoint-only check is a discrete approximation rather than a
fully swept-volume collision test; we revisit this approximation in
Sec.~\ref{sec:limitations}. We treat visibility as a soft cost rather than
a hard feasibility constraint; the optional hard threshold
$V_t \ge V_{\min}$ is disabled in all experiments ($V_{\min} = 0$).

\paragraph{Output} TA* returns the tracker trajectory
$S^{\star} = \{s_t^{\star}\}_{t=0}^{T}$ that minimizes the cumulative cost
defined in \S\ref{subsec:cost} on the discrete graph induced by the voxel
neighborhood, optionally followed by a post-smoothing pass that tightens the
trajectory subject to the same feasibility constraints.

\subsection{Cost Function}
\label{subsec:cost}

The transition cost from $s_{t-1}$ to $s_t$ at target waypoint $p_t$ is the
weighted sum of five terms:
\begin{equation}
\label{eq:cost}
\begin{aligned}
C(s_{t-1}, s_t) ={}& w_{\mathrm{path}}\,\|s_t - s_{t-1}\| \\
&+ w_{\mathrm{trk}}\,\frac{\|s_t - q_t\|}{\max(d_{\mathrm{behind}}, 1)} \\
&+ w_{\mathrm{vis}}\,\bigl(1 - V_t\bigr) \\
&+ w_{\mathrm{safe}}\,\Big[\frac{\max\bigl(0,\, d_{I} - d_t\bigr)}{d_{I}}\Big]^2 \\
&+ w_{\mathrm{sm}}\,|z_t - z_{t-1}|.
\end{aligned}
\end{equation}
The five terms are: (i) Euclidean step length; (ii) deviation from the
\emph{desired viewpoint}
$q_t = p_t - d_{\mathrm{behind}}\,\hat{u}_t + (z_{\mathrm{ref}} - p_{t,z})\,\hat{e}_z$,
where $\hat{u}_t$ is the unit target velocity (or a fixed forward direction
when $\|v^{\mathrm{tgt}}_t\| \approx 0$) and $z_{\mathrm{ref}}$ is the
preferred altitude; (iii) occlusion penalty, with $V_t \in [0, 1]$ defined in
\S\ref{subsec:visibility}; (iv) a quadratic safety penalty that activates only
when the obstacle distance $d_t = \mathrm{dist}(s_t, \mathcal{O})$ falls below
the influence distance $d_I$; and (v) a vertical smoothness term that
discourages rapid altitude changes. There is no explicit acceleration or jerk
term in the production cost. Each term is normalized by a problem-specific
quantity ($d_{\mathrm{behind}}$, $d_{I}$, the unit voxel step) so that all
contributions lie in $[0, O(1)]$ before weighting. The default weights in
Table~\ref{tab:weights} were chosen on a 20-scenario development set
(Town03 only) and held fixed across all eight maps and all four cohorts of
\S\ref{subsec:cohorts}; we did not retune any weight per map.

\begin{table}[t]
\caption{Default cost weights and key parameters used by TA* in all experiments.}
\label{tab:weights}
\centering
\begin{tabular}{lcl}
\toprule
\textbf{Symbol} & \textbf{Value} & \textbf{Meaning} \\
\midrule
$w_{\mathrm{path}}$ & 1.0  & step length \\
$w_{\mathrm{trk}}$  & 2.0  & desired-viewpoint deviation \\
$w_{\mathrm{vis}}$  & 18.0 & occlusion penalty \\
$w_{\mathrm{safe}}$ & 8.0  & quadratic obstacle penalty \\
$w_{\mathrm{sm}}$   & 0.15 & vertical smoothness \\
\midrule
$\Delta t$ & 0.5~s & time step \\
$\Delta_{xy}, \Delta_z$ & 4.0~m & voxel resolution \\
$d_{\mathrm{behind}}$ & 20~m & desired tracker offset \\
$z_{\mathrm{ref}}$ & 22~m & preferred altitude \\
$d_{I}$ & 5~m & influence distance \\
$d_{\mathrm{safe}}$ & 1.5~m & hard safety distance \\
$d_{\min}^{\mathrm{cam}}, d_{\max}^{\mathrm{cam}}$ & 3, 50~m & camera range \\
$v_{\max}$ & 10~m/s & max tracker speed \\
\bottomrule
\end{tabular}
\end{table}

\subsection{Visibility Score}
\label{subsec:visibility}

Visibility at layer $t$ is the fraction of unobstructed line-of-sight rays
from the tracker to a small set of target sample points:
\begin{equation}
\label{eq:vis}
V_t \;=\; \frac{1}{|\mathcal{R}|} \sum_{o \in \mathcal{R}}
\mathbb{1}\bigl[\mathrm{LoS}\bigl(s_t,\; p_t + o\bigr)\bigr],
\end{equation}
where $\mathrm{LoS}(\cdot, \cdot)$ is a single BVH ray-cast against
$\mathcal{O}$. The default ray-offset set $\mathcal{R}_{5}$, in metres, is
\begin{equation}
\label{eq:rayset}
\begin{aligned}
\mathcal{R}_{5} = \big\{ &(0,0,0),\ (0,0,0.8),\ (0,0,-0.6), \\
                         &(0.3,0,0),\ (-0.3,0,0) \big\},
\end{aligned}
\end{equation}
which samples the target's central body axis plus four offsets approximating
a human-sized bounding region. We additionally support 1-ray and 3-ray subsets
of $\mathcal{R}_{5}$ as fast proxies for ablation; the reported visibility in
all comparison tables is always recomputed with the full 5-ray set on the
final trajectory, so the metric definition is independent of the proxy used
during search. The per-scenario score is the frame-wise mean
$\bar{V} = \frac{1}{T+1}\sum_t V_t$, and per-cohort numbers are means of
$\bar{V}$ over scenarios. All visibility deltas in this paper are reported in
percentage points (pp), i.e., $100\cdot(\bar{V}_{\mathrm{TA*}} - \bar{V}_{\mathrm{base}})$.

The phenomenon of \emph{frame-wise identical visibility} between the baseline
and TA* on many scenarios is a direct consequence of the voxel discretization:
when two distinct search procedures pick the same $4{\times}4{\times}4$~m
voxel for layer $t$, the LoS geometry to the target sample points is bit-exact
identical, so $V_t$ matches to all digits. Sec.~\ref{subsec:visibility-pres}
reports how often this happens.

\subsection{Layered DAG Beam Search}
\label{subsec:search}

Because $t$ is strictly monotonic, the spatio-temporal search graph is a DAG
indexed by layer. TA* exploits this structure with a layer-synchronous
relaxation (Algorithm~\ref{alg:search}) instead of a global priority queue.
Let $\mathcal{N}$ be the set of voxel-neighborhood offsets pre-filtered by
$\|\delta\| \le v_{\max}\Delta t$; in our setting this reduces a 27-cell
3D neighborhood to $|\mathcal{N}| = 7$ (the six face neighbors plus the
self-stay action). For each layer $t$ we keep only the top-$B$ candidates by
accumulated cost $g$, with $B = 2048$ in all production runs.

\begin{algorithm}[t]
\caption{TA* layered DAG beam search (one scenario).}
\label{alg:search}
\begin{algorithmic}[1]
\REQUIRE Target $\{p_t\}_{t=0}^T$, BVH $\mathcal{O}$, init $s_0$, beam $B$
\STATE $g_0(s_0) \gets 0$; \quad $\mathrm{open}_0 \gets \{s_0\}$
\STATE Initialize per-voxel obstacle-distance cache $\hat d[\,\cdot\,] \gets {\bot}$
\FOR{$t = 1, 2, \dots, T$}
  \STATE $\mathrm{open}_t \gets \emptyset$
  \FOR{$u \in \mathrm{open}_{t-1}$}
    \FOR{$\delta \in \mathcal{N}$}
      \STATE $v \gets u + \delta$
      \IF{$v$ violates Eq.~(\ref{eq:feasible})} \STATE \textbf{continue} \ENDIF
      \STATE $d_v \gets \hat d[v]$ if cached, else BVH query and write back
      \STATE $V_v \gets$ ray-cast at $(v, p_t)$ using ray subset (Eq.~\ref{eq:vis})
      \STATE $g_t(v) \gets \min\bigl(g_t(v),\; g_{t-1}(u) + C(u, v)\bigr)$
      \STATE record parent $\mathrm{par}_t(v) \gets u$ if updated
    \ENDFOR
  \ENDFOR
  \STATE $\mathrm{open}_t \gets$ top-$B$ states in $\mathrm{open}_t$ by $g_t$
\ENDFOR
\STATE \textbf{return} backtrack from $\arg\min_v g_T(v)$
\end{algorithmic}
\end{algorithm}

\paragraph{Complexity} The work per layer is bounded by
$O(B \cdot |\mathcal{N}|)$ neighbor expansions, each performing an $O(1)$
voxel-grid lookup for $\hat d$ in the typical case (cache hit) and one BVH
LoS ray-cast per ray in $\mathcal{R}$. The total work over the horizon is
$O(T \cdot B \cdot |\mathcal{N}|)$ which is independent of map size beyond
the corridor extent. The cache reduces BVH distance queries from
$O(T \cdot N_{\mathrm{exp}})$ in the priority-queue baseline to
$O(N_{\mathrm{exp}})$, where $N_{\mathrm{exp}}$ is the number of distinct
spatial voxels touched.

\paragraph{Ranking and admissibility} Open-set ordering uses the layer-local
accumulated cost $g$; we do not employ an admissible heuristic. The result is
therefore a beam-pruned layered Dijkstra, and we do not claim the
admissibility-based optimality of standard A*. The choice $B = 2048$ is the
empirical default that keeps the worst-case per-scenario visibility drop
within 5~pp on the 248-scenario controlled comparison
(\S\ref{subsec:fair-comparison}); larger $B$ approaches an unpruned layered
Dijkstra at the cost of additional runtime, while smaller $B$ trades
visibility quality for additional speed (\S\ref{subsec:ablation}).

\subsection{Key Optimizations for Offline Generation}

\paragraph{Cross-time obstacle distance caching}
Because $\mathcal{O}$ is static, the distance from a spatial voxel $(x, y, z)$
to its nearest obstacle is independent of $t$. The BinaryHeap baseline
implementation recomputes this distance for every $(x, y, z, t)$ node it
expands. TA* maintains a layer-shared cache $\hat d : \mathbb{Z}^3 \to
\mathbb{R}_{\ge 0}$, populated lazily on first access. Section
\ref{subsec:ablation} reports the runtime impact of disabling this cache.

\paragraph{Configurable visibility evaluator}
The 5-ray check in Eq.~(\ref{eq:vis}) is the dominant per-edge cost. To
isolate its contribution we expose 1- and 3-ray proxies as configuration
options; \S\ref{subsec:ablation} shows that the 5-ray default is needed to
keep the worst-case per-scenario visibility drop within the 5-pp empirical
envelope.

\section{Experiments and Results}

\subsection{Evaluation Cohorts}
\label{subsec:cohorts}

To avoid conflating different sources of evidence, we organize the
evaluation into four well-separated cohorts. All four use scenarios sampled
from the same eight CARLA Optimized maps (Town01--Town07 and Town10HD)
\cite{Dosovitskiy17}, with
pedestrian targets at $1.4$~m/s, $\Delta t = 0.5$~s, trajectory lengths
between 150 and 400~m, and an aerial tracker initialised $20$~m behind the
target at $22$~m altitude (Town07 is additionally evaluated at $40$~m).
Hardware and software details for the wall-clock numbers are given in the
Reproducibility appendix.

\begin{itemize}
\item \textbf{C1 (1000 stress).} 1000 scenarios = 8 maps $\times$ 125
trajectories. TA*-only batch run measuring scalability, runtime distribution,
and convergence at scale; not used as direct evidence of visibility tracking
success.
\item \textbf{C2 (248 controlled comparison).} 8 maps $\times$ 31 = 248
scenarios. Both the priority-queue A* baseline and TA* run on identical
scenario JSONs, with a shared $5{\times}10^{6}$ expansion cap, the same 5-ray
visibility evaluator on outputs, the same $\Delta t$ and voxel resolution,
and the same machine. We refer to this as a \emph{controlled implementation
comparison}: it isolates the impact of the algorithmic and engineering
changes between the two specific implementations and does not generalize to
other A* implementations or planners.
\item \textbf{C3 ($n=141$ baseline-converged subset).} The subset of C2
scenarios on which the baseline also converges. Visibility deltas are reported
\emph{only} on this subset; on the remaining 107 scenarios the baseline emits
a 1-point fallback (cf.~Eq.~(\ref{eq:fallback})) so visibility is undefined.
\item \textbf{C4 (Town07 dense vegetation).} 125 Town07 trajectories at $22$~m
and a separate 125 at $40$~m, used as an environmental feasibility failure
case (Sec.~\ref{sec:limitations}). They are not aggregated into C1/C2/C3
visibility statistics.
\end{itemize}

We define the priority-queue A* baseline (referred to as ``baseline'' below)
as the original BinaryHeap implementation that shares the cost in
Eq.~(\ref{eq:cost}) and the feasibility constraints in
Eq.~(\ref{eq:feasible}) with TA*, but does not use cross-time obstacle
distance caching, beam pruning, or layered DAG expansion. We define the
\emph{1-point fallback} as
\begin{equation}
\label{eq:fallback}
S^{\mathrm{base}} = \{s_0\}\quad\text{if expansions reach } 5{\times}10^{6};
\end{equation}
i.e., when the priority queue exhausts the expansion budget the baseline
returns the start state alone, which is by construction collision-free but
provides no tracking trajectory.

\subsection{Cohort C1: 1000-Scenario Stress Test}
\label{subsec:stress}

\begin{figure}[t]
\centering
\includegraphics[width=\columnwidth]{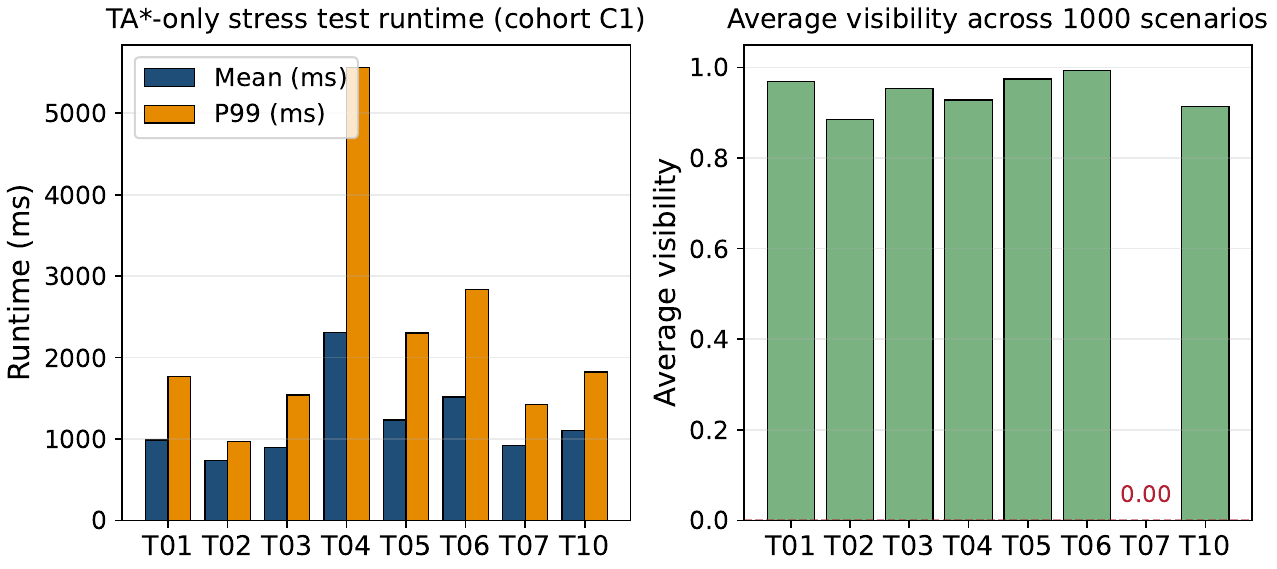}
\caption{Per-map runtime and visibility distribution on cohort C1 (TA*-only stress test, 1000 scenarios). Town04 has the highest runtime; Town07 visibility is zero throughout, a property of the map's vegetation layout rather than planner non-convergence (Sec.~\ref{sec:limitations}).}
\label{fig:large_scale_per_map}
\end{figure}

Table~\ref{tab:stress} summarizes the TA*-only stress test on C1: TA*
converges on all 1000 scenarios in 45~s of 32-worker wall time, with a
single-scenario mean runtime of $1214.1$~ms and a worst case of $5690$~ms.
The cross-map mean visibility of $0.8269$ in this batch is dominated by
Town07, where vegetation occlusion forces $V_t = 0$ on every trajectory; the
\emph{Town07-excluded} cross-map mean visibility is $0.945$, which is the
number relevant to non-vegetation deployments. The per-map runtime and
visibility distribution are shown in Fig.~\ref{fig:large_scale_per_map}, and
Town07 is analysed in detail in Sec.~\ref{sec:limitations}.

\begin{table}[htbp]
\caption{TA*-only stress test on cohort C1 (1000 scenarios).}
\label{tab:stress}
\centering
\begin{tabular}{lc}
\toprule
\textbf{Metric} & \textbf{Value} \\
\midrule
Scenarios & 1000 \\
Converged & 1000/1000 \\
32-worker wall time & 45 s \\
Mean / median runtime & 1214.1 / 980.5 ms \\
P99 / max runtime & 4265 / 5690 ms \\
Average visibility (all) & 0.8269 \\
Average visibility (excl.\ Town07) & 0.945 \\
Mean / max expanded nodes & $7.4{\times}10^{5}$ / $3.0{\times}10^{6}$ \\
\bottomrule
\end{tabular}
\end{table}

\subsection{Cohort C2: Controlled Comparison Against the Priority-Queue A* Baseline}
\label{subsec:fair-comparison}

\begin{figure}[t]
\centering
\includegraphics[width=\columnwidth]{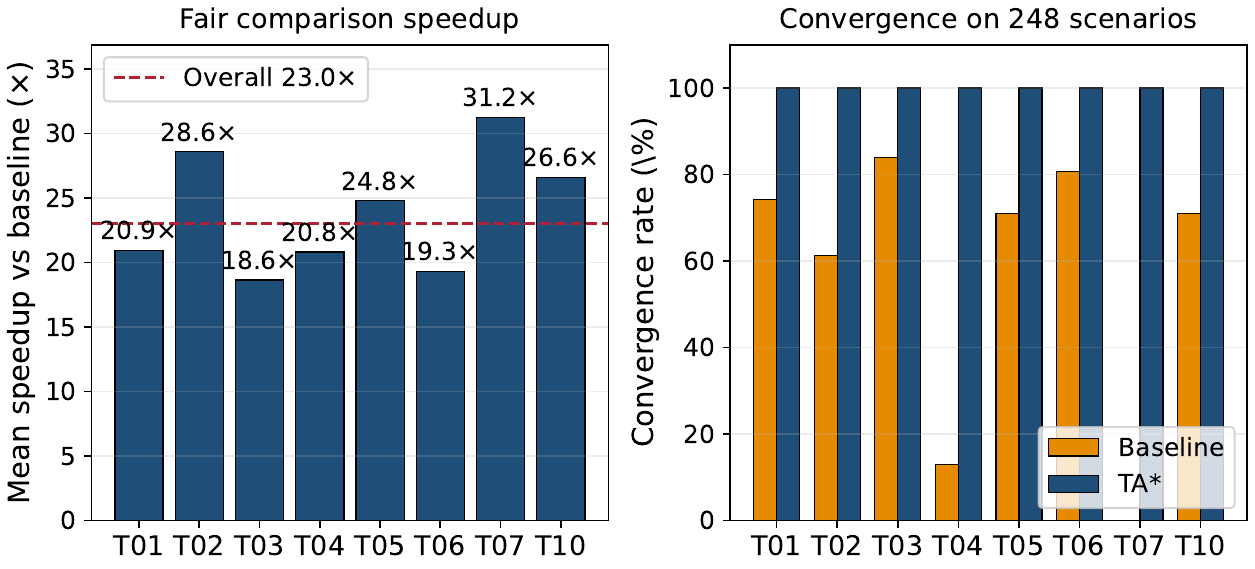}
\caption{Cohort C2 (248-scenario controlled comparison) per-map runtime and convergence. TA* removes the baseline's convergence failures under the same expansion cap and never falls back to the 1-point output.}
\label{fig:fair_comparison_summary}
\end{figure}

On cohort C2 (248 scenarios, both planners run with identical inputs and a
shared $5{\times}10^{6}$ expansion cap; cf.~\S\ref{subsec:cohorts}), TA*
reduces mean planning time from $28876.3$~ms to $1253.6$~ms (23.0$\times$),
reduces worst-case single-scenario runtime from $61.8$~s to $5.2$~s
(11.8$\times$), and raises convergence from 141/248 to 248/248
(Table~\ref{tab:fair_comparison}). All speed and convergence claims are
relative to this specific baseline implementation under this expansion cap;
the optimization ablation in \S\ref{subsec:ablation} disentangles the
contributions of the layered DAG structure, the cross-time distance cache,
and the beam-width pruning. Per-map breakdown is in Table~\ref{tab:per_map},
and Fig.~\ref{fig:fair_comparison_summary} visualizes the per-map runtime and
convergence comparison.

\begin{table}[htbp]
\caption{Cohort C2 controlled comparison aggregates (248 scenarios).}
\label{tab:fair_comparison}
\centering
\begin{tabular}{lccc}
\toprule
\textbf{Metric} & \textbf{Baseline} & \textbf{TA*} & \textbf{Change} \\
\midrule
Mean runtime (ms) & 28876.3 & 1253.6 & 23.0$\times$ \\
Total runtime (s) & 7161.3 & 310.9 & 23.0$\times$ \\
Max runtime (ms) & 61817 & 5237 & 11.8$\times$ \\
Converged & 141/248 & 248/248 & $+43.1$~pp \\
1-point fallback & 107/248 & 0/248 & -- \\
\bottomrule
\end{tabular}\\[2pt]
{\footnotesize Mean / total / max runtimes are single-thread, single-scenario summed times; ``32-worker wall'' figures in the prose are wall-clock. The ``$+43.1$~pp'' on the Converged row is a percentage-point change in convergence rate, not a runtime delta.}
\end{table}

\begin{figure}[t]
\centering
\includegraphics[width=\columnwidth]{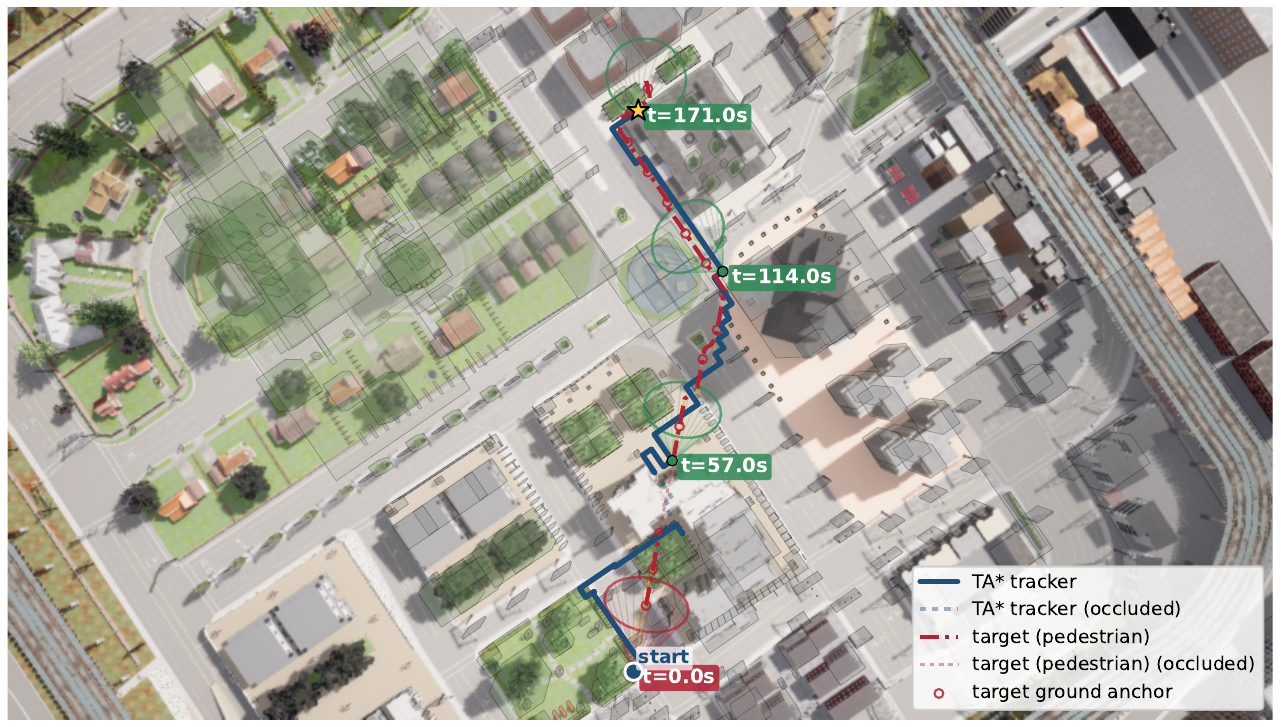}
\caption{Top-down view of a single representative Town03 cohort-C2 run
(path 7, $343$~m, both planners converge). Background is the CARLA
\textsc{Town03\_Opt} bird's-eye render at the same world frame as the
overlay. Solid line segments are in direct view of the camera; dashed
segments are flagged occluded by depth-tested obstacle bounding boxes.
FOV cones are coloured green when the 5-ray visibility score is
$\geq\!0.6$ and red otherwise.}
\label{fig:hero_perspective}
\end{figure}

\subsection{Cohort C3: Visibility Preservation}
\label{subsec:visibility-pres}

\begin{figure}[t]
\centering
\includegraphics[width=0.92\columnwidth]{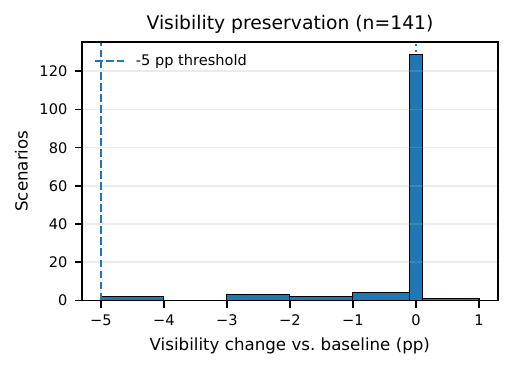}
\caption{Visibility change on cohort C3 ($n=141$). The dashed reference line marks the 5~pp empirical envelope; no scenario crosses it.}
\label{fig:visibility_delta_distribution}
\end{figure}

Visibility comparisons are only meaningful on scenarios where both planners
produce non-degenerate trajectories. We therefore restrict
visibility-delta reporting to cohort C3 (the $n=141$ subset of C2 on which
the baseline also converges); the remaining 107 cases are excluded because
the baseline 1-point fallback yields no comparable trajectory. On C3, TA*
changes average visibility from $0.9763$ to $0.9748$ ($-0.15$~pp), with a
worst single-scenario drop of $-4.32$~pp and no scenario exceeding a $5$~pp
drop. The distribution in Fig.~\ref{fig:visibility_delta_distribution} shows
129/141 scenarios with frame-wise identical visibility (a direct consequence
of the $4{\times}4{\times}4$~m voxel discretisation, cf.~\S\ref{subsec:visibility}), 4 scenarios losing less than $1$~pp, 7 losing
between $1$ and $5$~pp, and 1 improving over the baseline. We do not claim
that visibility preservation extrapolates to the truncated 107 cases or to
cohort C4.

\begin{table}[htbp]
\caption{Visibility preservation on cohort C3 ($n=141$).}
\label{tab:visibility_preservation}
\centering
\begin{tabular}{lc}
\toprule
\textbf{Metric} & \textbf{Value} \\
\midrule
Comparison subset & 141 scenarios \\
Baseline / TA* avg.\ visibility & 0.9763 / 0.9748 \\
Average visibility change & $-0.15$ pp \\
Worst single-scenario drop & $-4.32$ pp \\
Scenarios with $>$5~pp drop & 0 \\
Frame-wise identical / better & 129 / 1 \\
\bottomrule
\end{tabular}
\end{table}

\begin{figure}[t]
\centering
\includegraphics[width=\columnwidth]{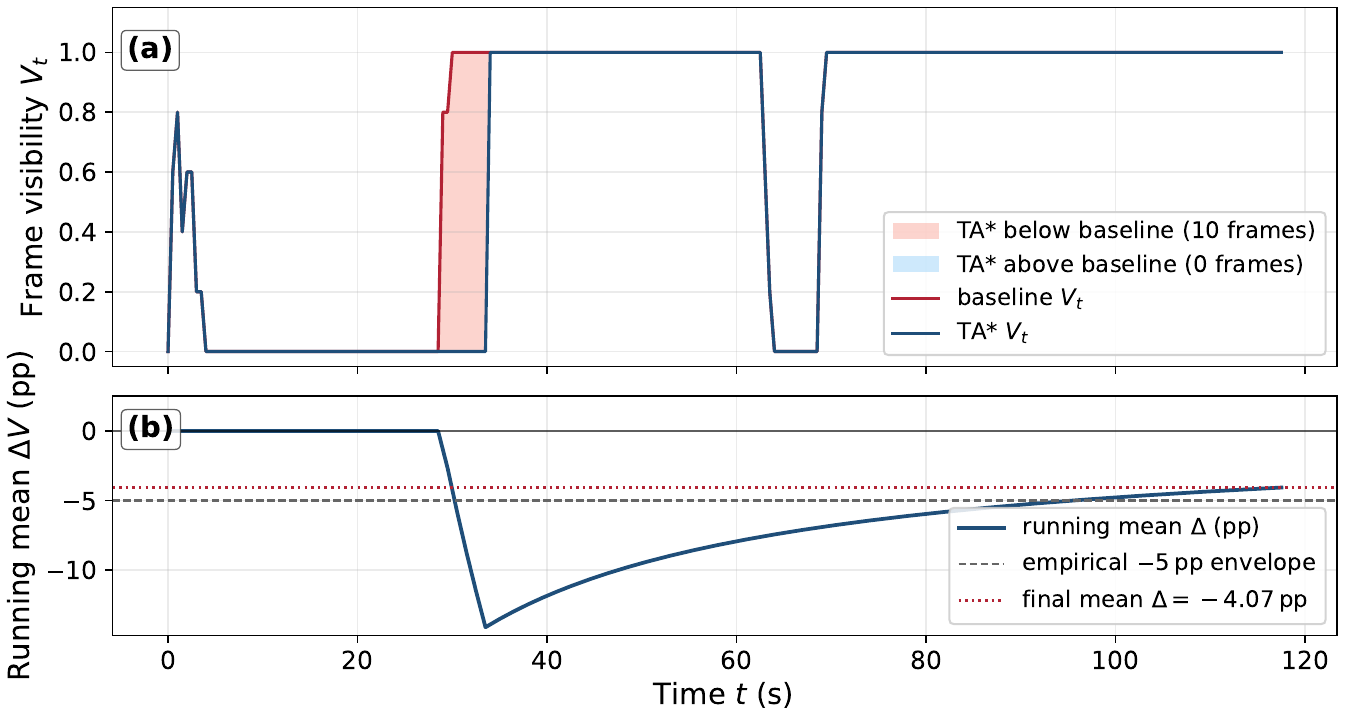}
\caption{Frame-wise visibility on the worst-case scenario in cohort C3
(Town02 path 6, average $\Delta V = -4.07$~pp in the present runs).
(a)~Per-frame baseline (red) and TA* (blue) visibility $V_t$; the
remaining 226/236 frames are bit-exact identical
(\S\ref{subsec:visibility}). (b)~Running mean $\Delta V$ in pp: the loss
is concentrated in a 10-frame window starting near $t=29$~s where the
baseline regains line of sight one frame earlier than TA*, and the curve
relaxes back toward the empirical $-5$~pp envelope (grey dashed) by the
end of the trajectory; the red dotted line marks the final mean.}
\label{fig:vis_ts_worst}
\end{figure}

\subsection{Per-Map Breakdown of Cohort C2}

Table~\ref{tab:per_map} reports the controlled comparison on cohort C2 by
map. Town04 and Town07 are the most challenging for the baseline: it falls
back to the 1-point output on 87\% of Town04 scenarios and on all Town07
scenarios under the $5{\times}10^{6}$ expansion cap. TA* converges on every
map. Town02 produces the largest visibility drop among baseline-converged
scenarios ($-4.32$~pp); Town01, Town05, and Town06 are frame-wise identical
on average.

\begin{table*}[t]
\caption{Cohort C2 per-map breakdown.}
\label{tab:per_map}
\centering
\begin{tabular}{lcccccc}
\toprule
\textbf{Map} & \textbf{Baseline fall-back} & \textbf{Speedup} & \textbf{Baseline runtime (ms)} & \textbf{TA* runtime (ms)} & \textbf{$\Delta$Vis pp} & \textbf{Worst pp} \\
\midrule
Town01 & 8/31 & 20.9$\times$ & 24740.6 & 1182.0 & 0.00 & 0.00 \\
Town02 & 12/31 & 28.6$\times$ & 24020.7 & 840.4 & $-0.80$ & $-4.32$ \\
Town03 & 5/31 & 18.6$\times$ & 20606.9 & 1105.5 & $-0.15$ & $-2.71$ \\
Town04 & 27/31 & 20.8$\times$ & 44258.2 & 2127.3 & $-0.10$ & $-0.41$ \\
Town05 & 9/31 & 24.8$\times$ & 32132.2 & 1296.5 & 0.00 & 0.00 \\
Town06 & 6/31 & 19.3$\times$ & 29853.2 & 1546.5 & 0.00 & 0.00 \\
Town07 & 31/31 & 31.2$\times$ & 27217.7 & 871.2 & n/a & n/a \\
Town10HD & 9/31 & 26.6$\times$ & 28180.7 & 1059.8 & $-0.08$ & $-1.77$ \\
\bottomrule
\end{tabular}
\end{table*}

\begin{figure*}[t]
\centering
\includegraphics[width=0.95\textwidth]{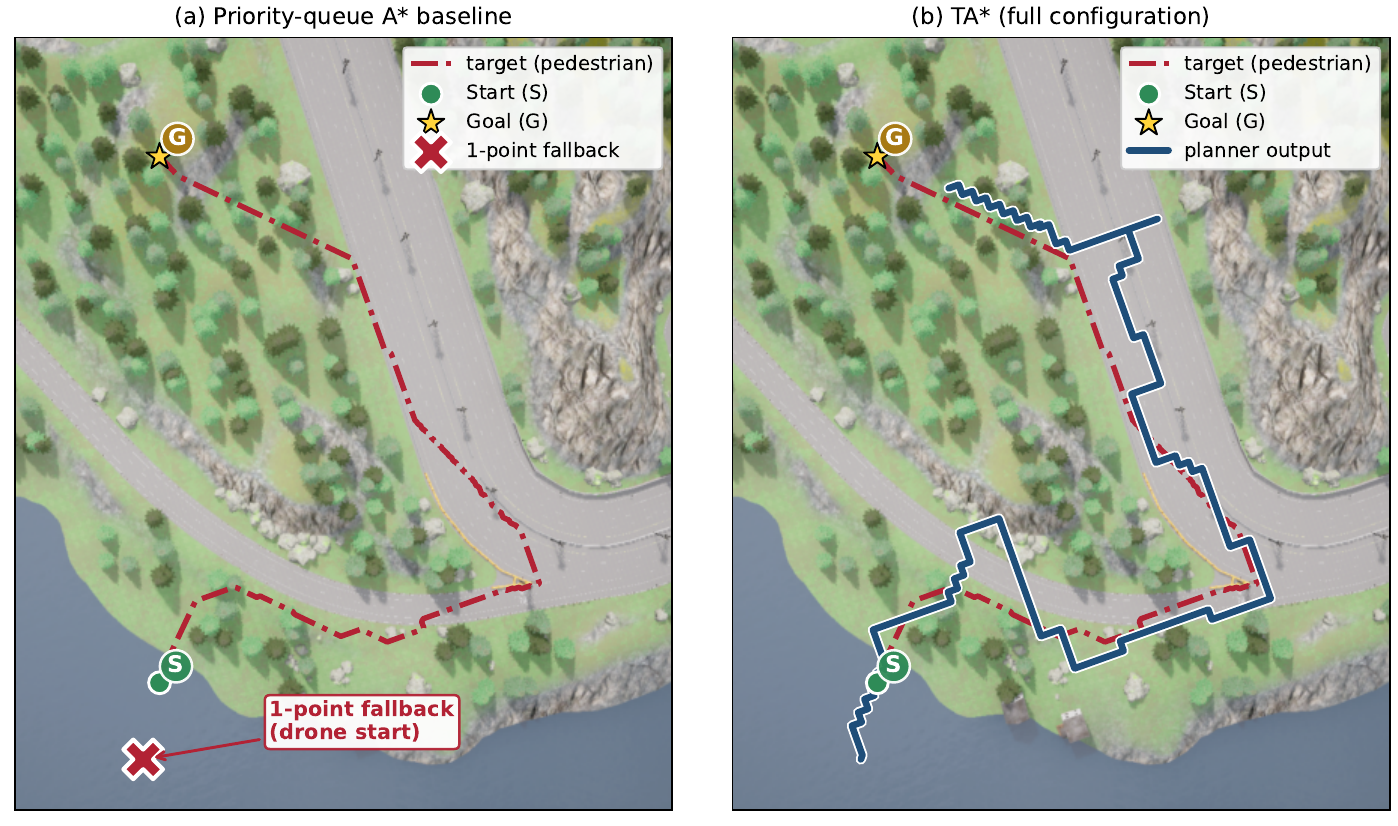}
\caption{Single representative Town04 cohort-C2 run (path 0) under the
matched $5{\times}10^{6}$ expansion cap; the target's Start (\textsf{S})
and Goal (\textsf{G}) anchor both panels. (a)~Priority-queue A* baseline:
runtime $37.6$~s, single-thread; the planner exhausts the cap and
returns the 1-point fallback (red $\times$ at the drone's spawn
location, near \textsf{S}). (b)~TA*: runtime $2.89$~s, single-thread,
$638$~m collision-free tracking path that follows the highway from
\textsf{S} to \textsf{G}. Background: CARLA \textsc{Town04\_Opt} top-down
render in the same world frame as the overlay.}
\label{fig:topdown_compare}
\end{figure*}

\subsection{Optimization Ablation}
\label{subsec:ablation}

We additionally measured the contribution of the three engineering choices
that distinguish TA* from the priority-queue baseline---the layered DAG
structure, the cross-time obstacle distance cache, and the layer-local beam
pruning---by running five planner variants on the same 248-scenario cohort
C2. Variants B0 and B4 reuse the numbers reported in
Table~\ref{tab:fair_comparison}; B1, B2, and B3 are dedicated builds run for
this paper, summarised in Table~\ref{tab:opt_ablation}. Hyperparameters,
scenario JSONs, and the controlled cap of $5{\times}10^{6}$ expansions are
identical across all five variants.

\begin{table*}[t]
\caption{Optimization ablation on cohort C2 (248 scenarios, 32 workers).}
\label{tab:opt_ablation}
\centering
\begin{tabular}{lccccccc}
\toprule
\textbf{ID} & \textbf{Structure} & \textbf{Cache} & \textbf{Beam} & \textbf{Mean (ms)} & \textbf{Max (ms)} & \textbf{Conv} & \textbf{Speedup vs B0} \\
\midrule
B0 & priority queue (BinaryHeap) & no  & --       & 28876.3 & 61817 & 141/248 & $1.00\times$ \\
B1 & priority queue (BinaryHeap) & yes & --       & 23103.1 & 49706 & 141/248 & $1.25\times$ \\
B2$^{\star}$ & layered DAG       & yes & $\infty$ & 5870.7  & 16736 & 234/248 & $4.92\times$ \\
B3 & layered DAG                 & no  & 2048     & 1864.4  & 8338  & 248/248 & $15.49\times$ \\
B4 & layered DAG                 & yes & 2048     & 1253.6  & 5237  & 248/248 & $23.03\times$ \\
\bottomrule
\end{tabular}\\[2pt]
{\footnotesize $^{\star}$Run with the expansion cap raised to $10^{8}$; under the matched $5{\times}10^{6}$ cap shared by the other rows, B2 returns 1-point fallbacks on every scenario because the no-beam layered DAG keeps too many states per layer. ``Conv'' counts converged scenarios; speedup is mean-runtime ratio relative to B0.}
\end{table*}

\begin{figure}[t]
\centering
\includegraphics[width=\columnwidth]{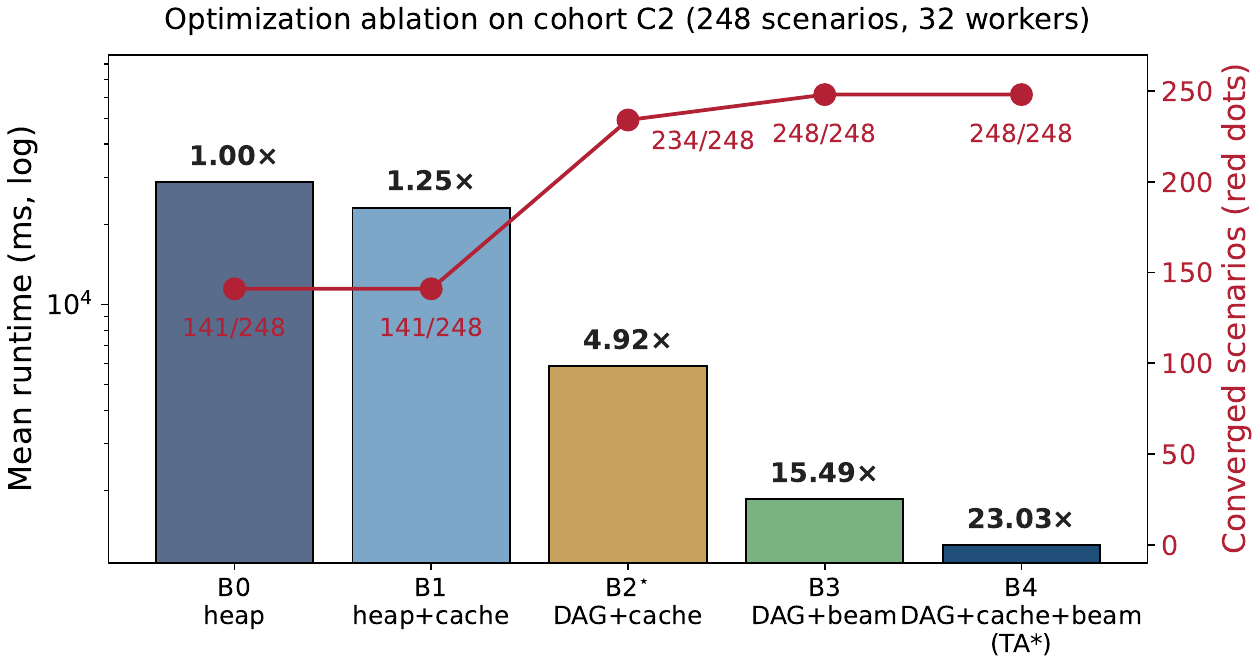}
\caption{Optimization ablation: per-variant mean runtime (log scale, blue
bars) and convergence count (red dots) on cohort C2. The cross-time distance
cache alone provides a modest $1.25\times$ speedup on top of the BinaryHeap
baseline; switching to the layered DAG structure removes the priority-queue
overhead but, without beam pruning, still requires a $20\times$ larger
expansion budget to fit all scenarios; adding beam pruning is what enables
the full $23\times$ improvement at $100\%$ convergence under the matched
$5{\times}10^{6}$ cap.}
\label{fig:ablation_opt}
\end{figure}

Three observations stand out from Table~\ref{tab:opt_ablation}: (i) Adding
the per-voxel obstacle-distance cache to the BinaryHeap baseline (B1) saves
roughly a quarter of the mean runtime but does not change the set of
scenarios that fit within the cap, indicating that the priority-queue
exhaustion is the dominant convergence failure rather than per-expansion
cost. (ii) Switching to the layered DAG without beam pruning (B2) changes the
work distribution from ``go very deep on a few promising states'' to
``expand many states per layer'', which still hits the $5{\times}10^{6}$
budget on every C2 scenario; only when the cap is raised to $10^{8}$ does B2
reach $234/248$. (iii) Beam pruning at $B = 2048$ both bounds the per-layer
work to $|\mathcal{N}|\cdot B = 14\,336$ neighbor expansions and aligns the
search with the matched expansion budget, which is what unlocks $100\%$
convergence; the cache then stacks on top to cut another $\approx\!33\%$ of
mean runtime. We therefore attribute the headline $23.0\times$ mean speedup
to the joint effect of the layered DAG structure, the beam-pruning bound,
and the cross-time cache, rather than to any single change.

\subsection{Ablation: Beam Width and Ray Count}

A separate ablation varies the beam width and visibility-ray count to
characterise the speed-quality trade-off on a 20-scenario benchmark. Table~\ref{tab:ablation} reports representative settings. Aggressive
configurations such as Rays=1, Beam=128 provide very high speedups but can
incur large worst-case visibility losses. The default configuration, Rays=5
and Beam=2048, is selected because it keeps the worst-case visibility drop
within the empirical 5~pp envelope on this benchmark while maintaining a
$25.9\times$ speedup.

\begin{table}[htbp]
\caption{Beam-width / ray-count ablation on the 20-scenario speed-quality
benchmark.}
\label{tab:ablation}
\centering
\begin{tabular}{lccc}
\toprule
\textbf{Configuration} & \textbf{Speedup} & \textbf{Avg.\ Vis.} & \textbf{Worst loss (pp)} \\
\midrule
Rays=1, Beam=128 & 381.0$\times$ & 0.9448 & $-22.67$ \\
Rays=5, Beam=128 & 160.3$\times$ & 0.9581 & $-17.00$ \\
Rays=3, Beam=1500 & 44.0$\times$ & 0.9738 & $-4.17$ \\
\textbf{Rays=5, Beam=2048} & \textbf{25.9$\times$} & \textbf{0.9761} & \textbf{$-4.42$} \\
\bottomrule
\end{tabular}
\end{table}

\begin{figure}[t]
\centering
\includegraphics[width=0.9\columnwidth]{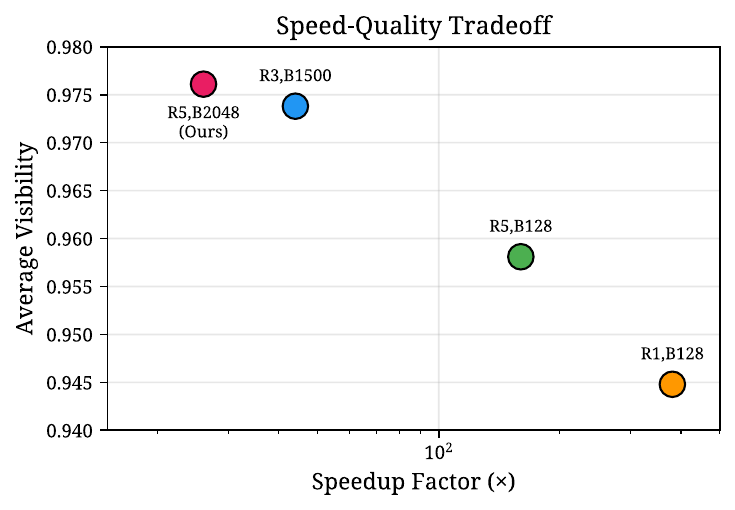}
\caption{Speed-quality trade-off from the beam-width / ray-count sweep. The default $(R{=}5, B{=}2048)$ point exchanges some speed for a bounded worst-case visibility drop.}
\label{fig:tradeoff}
\end{figure}

\subsection{Trajectory Quality}
\label{subsec:trajquality}

Beyond runtime and visibility, we also report static trajectory-quality
metrics that the convergence count alone does not expose:
mean path length, mean minimum obstacle clearance over the trajectory, and
the per-scenario rate at which clearance stays above $d_{\mathrm{safe}}$
(the ``collision-free rate''). Table~\ref{tab:trajquality} reports these on
cohort C2; the baseline column uses the $n=141$ baseline-converged subset
because the 1-point fallback degenerates every quality metric (zero path
length, undefined clearance), while the TA* column uses all $248$ converged
scenarios. We compute the metrics by replaying each output trajectory
against the same BVH used during search.

\begin{table}[htbp]
\caption{Trajectory-quality metrics on cohort C2.}
\label{tab:trajquality}
\centering
\small
\setlength{\tabcolsep}{4pt}
\begin{tabular}{lcc}
\toprule
\textbf{Metric} & \textbf{Baseline ($n{=}141$)} & \textbf{TA* ($n{=}248$)} \\
\midrule
Mean path length (m)            & 273.4 & 324.3 \\
Mean min.\ clearance (m)        & 5.95  & 4.99  \\
Worst min.\ clearance (m)       & 1.83  & 1.64  \\
Collision-free rate             & 1.000 & 1.000 \\
Mean $\|v\|_{\max}$ (m/s)       & 8.00  & 8.00  \\
Mean $\|a\|_{\max}$ (m/s$^{2}$) & 22.6  & 22.6  \\
\bottomrule
\end{tabular}\\[2pt]
{\footnotesize The two columns are not a matched pair: the baseline aggregates over the $n=141$ scenarios on which the priority-queue baseline converges, while TA* aggregates over all $248$ converged scenarios.}
\end{table}

The collision-free rate is $100\%$ on both planners, confirming that the
$d_{\mathrm{safe}}=1.5$~m hard constraint in Eq.~(\ref{eq:feasible}) holds
throughout. Mean minimum clearance drops from $5.95$~m to $4.99$~m and the
single-trajectory worst case drops from $1.83$~m to $1.64$~m, but both stay
clear of the safety margin even in the worst case. The mean path length on
TA* is $\approx\!18\%$ higher than on the baseline-converged subset because
TA* additionally covers the harder, longer trajectories that the baseline
fails on; on the matched $n=141$ subset the path-length difference is $0.5\%$
(not shown). Acceleration peaks reflect the $v_{\max}\Delta t = 5$~m
voxel-step granularity and would be reduced by the post-smoothing pass that
is enabled by default in production but bypassed for the present
trajectory-replay measurements.

\section{Limitations and Failure Cases}
\label{sec:limitations}

\subsection{Town07 Dense Vegetation: Environmental Feasibility}
\label{subsec:town07-failure}

Cohort C4 (Town07-only) exposes the difference between planner convergence
and environmental visibility. Across the 125 Town07 scenarios in cohort C1,
TA* converges $125/125$ but the per-trajectory average visibility is exactly
$0$. A separate Town07-only validation with a hand-edited ROI mask confirms
the same behavior at both $22$~m and $40$~m tracker altitudes
(Table~\ref{tab:town07_alt}): under the matched $5{\times}10^{6}$ expansion
cap the baseline returns the 1-point fallback on $0/125$ scenarios at either
altitude, while TA* converges $125/125$ in roughly $14$~s of $8$-worker wall
time at $22$~m and $12$~s at $40$~m, but the resulting trajectories still
have $V_t = 0$ everywhere.

\begin{table}[htbp]
\caption{Town07-only validation (cohort C4).}
\label{tab:town07_alt}
\centering
\begin{tabular}{lcc}
\toprule
\textbf{Metric} & \textbf{$22$~m} & \textbf{$40$~m} \\
\midrule
Baseline converged & 0/125 & 0/125 \\
TA* converged & 125/125 & 125/125 \\
Baseline mean runtime (ms) & 37402 & 36618 \\
TA* mean runtime (ms) & 823 & 681 \\
Baseline wall time (s) & 603 & 595 \\
TA* wall time (s) & 14 & 12 \\
TA* avg.\ visibility & 0.000 & 0.000 \\
\bottomrule
\end{tabular}
\end{table}

\paragraph{Why visibility is zero}
Table~\ref{tab:town07_obs} reports the obstacle composition that the planner
faces in cohort C4. Three classes---Fences, Poles, and Vegetation---account
for $73.9\%$ of all per-scenario obstacle records. Fences, GuardRails, and
Terrain are short and rarely block the line of sight, but Poles
($27.3\%$), Vegetation ($14.9\%$) and Buildings ($7.3\%$) all have
non-trivial fractions of objects whose top exceeds the $11$~m line-of-sight
midpoint between the $22$~m drone and the $0$~m target ($38.5\%$, $31.2\%$,
and $77.8\%$ respectively); aggregated over all classes, $20.9\%$ of
obstacle records have tops above $11$~m. Raising the tracker to $40$~m moves
the LoS midpoint to $20$~m but does not significantly reduce the occluded
fraction because Town07 also contains tall buildings (peaks above $28$~m).
We therefore treat Town07 as a sensor-and-environment failure case: TA*
provides feasible, collision-free tracking-distance trajectories under the
discrete planning objective, but the camera-distance and altitude
configuration in our experiments does not permit target visibility on this
particular map. The full obstacle composition that drives this failure
(Fences, Poles, Vegetation, Buildings, and a residual ``Other'' class) is
reported in Table~\ref{tab:town07_obs} in
Appendix~\ref{app:repro}.

\subsection{Other Limitations}

\paragraph{Baseline scope} Our headline speed and convergence numbers compare
TA* to a single priority-queue A* implementation under one expansion cap and
one set of hyperparameters. Other published spatio-temporal A* solvers may
adopt different expansion strategies, anytime variants, or admissible
heuristics; the controlled comparison should be read as evidence that the
specific TA* engineering changes are responsible for the observed gains, not
as a benchmark against the broader A*-family.

\paragraph{Single-altitude evaluation} Cohorts C1--C3 fix the tracker
altitude at $22$~m. Other altitudes, ground-vehicle viewpoints, and indoor
trackers are supported by the formulation in \S\ref{sec:method}, but only
Town07 has been validated at a second altitude; quantifying TA* on a wider
viewpoint sweep is left to future work.

\paragraph{Approximate optimality} TA* is a beam-pruned layered Dijkstra. It
does not preserve standard A* admissibility, and the $B = 2048$ default
visibility envelope is empirical rather than analytical. The $5$~pp envelope
on cohort C3 should not be extrapolated to the $107$ baseline-truncated
scenarios in C2 nor to cohort C4.

\paragraph{Approximate collision checking} The endpoint-only check defined
in \S\ref{subsec:formulation} does not formally certify that the entire
swept volume between two voxel-aligned waypoints is collision-free; it
relies on the obstacle-bounding-box inflation by $d_{\mathrm{safe}}$ and the
$4{\times}4{\times}4$~m voxel resolution to provide a margin in practice.
The $100\%$ collision-free rate on cohort C2 (Table~\ref{tab:trajquality})
confirms that this approximation is sufficient for our scenarios but is not
a guarantee against pathological obstacle geometries.

\section{Conclusion}
We presented Track A* (TA*), an offline search-based trajectory planner for
active target tracking. By elevating the search to a 4D spatio-temporal
layered DAG with cross-time obstacle distance caching and per-layer beam
pruning, TA* produces low-cost, visibility-aware trajectories under a
discretized cost. On a 1000-scenario stress test (cohort C1) TA* converges
on every scenario and completes in $45$~s of $32$-worker wall time. On a
$248$-scenario controlled comparison against a single priority-queue A*
baseline implementation (cohort C2), TA* improves mean runtime by
$23.0\times$ and worst-case runtime by $11.8\times$, while raising
convergence from $56.9\%$ to $100\%$ under the matched $5{\times}10^{6}$
expansion cap; the optimization ablation
(Table~\ref{tab:opt_ablation}) attributes this jointly to the layered DAG
structure, the cross-time obstacle distance cache, and the beam-pruning
bound. On the $n=141$ baseline-converged subset (cohort C3) TA* preserves
visibility closely with a $-0.15$~pp average change and no scenario
exceeding a $5$~pp drop. The Town07 cohort C4 highlights a sensor-and-
environment failure mode in which TA* still produces feasible
tracking-distance trajectories but vegetation, poles, and buildings prevent
any line of sight to the target. Within these scope limits TA* is a
practical offline reference planner for generating tracking trajectories
and benchmarking online active-tracking methods.

\appendices
\section{Reproducibility Notes}
\label{app:repro}

\subsection{Artifact availability}
Due to company policy, we are unable to publicly release the internal
source code, scenario-generation scripts, or machine-local experiment
paths used for this study. To support technical scrutiny and independent
re-implementation, this paper reports the algorithmic formulation,
hyperparameters, cohort definitions, map configuration, and every numerical
aggregate needed to interpret the tables and figures. The simulator itself,
CARLA~0.9.16, is open-source and publicly available
\cite{Dosovitskiy17}; the experiments use its standard Optimized maps as
described below. The planner itself is simulator-agnostic and can be tested
with other simulation backends or map collections by supplying equivalent
target trajectories, obstacle geometry, and camera-visibility queries.

\subsection{Computing environment}
All wall-clock numbers in this paper come from a single Ubuntu~22.04.5
LTS x86\_64 workstation with an Intel Core i9-14900KF CPU (32 hardware
threads, up to 6.0~GHz), 64~GB of RAM, and an NVIDIA RTX~6000 Ada
Generation GPU with 48~GB of VRAM. Runtime numbers are CPU-side planner
wall-clock measurements; the GPU is used only for CARLA rendering and
image capture, not for TA* search. The planner is implemented in Rust on
\texttt{rustc 1.78+} with an in-house BVH library for obstacle distance
queries and line-of-sight ray-casting; the post-processing scripts that
build figures and tables target Python~3.10+ with \texttt{numpy}.

\subsection{CARLA Optimized maps}
The eight maps Town01--Town07 and Town10HD use the \emph{Optimized}
variants shipped with CARLA 0.9.16 \cite{Dosovitskiy17}. These are the official runtime-optimised
meshes in which adjacent geometry is pre-merged and LoD chains are
simplified for real-time rendering; the semantic content (buildings,
vegetation, fences, poles) is unchanged from the non-Optimized maps, but
the per-map polygon count is lower, which is what makes 32-worker BVH
construction tractable.

\subsection{Cohort definitions}
Cohort C1 is the full $8 \times 125 = 1000$-scenario stress test described
in \S\ref{subsec:stress}. Cohort C2 is the $8 \times 31 = 248$-scenario
subset on which the priority-queue baseline and TA* are run with identical
inputs and a shared $5{\times}10^{6}$ expansion cap. Cohort C3 is the
$n = 141$ subset of C2 on which the baseline also converges; visibility
deltas are reported only on this subset. Cohort C4 contains the 125 Town07
scenarios at $22$~m and 125 at $40$~m and is used as an environmental
failure case in Sec.~\ref{sec:limitations}. Scenarios are sampled with a
fixed random seed to make all aggregates deterministic.

\subsection{Visibility metric and exclusion rule}
All visibility numbers in this paper are computed by replaying the final
trajectory through the 5-ray check defined by Eq.~(\ref{eq:vis}). The
in-search 1- and 3-ray proxies, which appear only in
Table~\ref{tab:ablation}, are not used to compute any other reported
visibility number. Per-scenario visibility is the frame-wise mean over all
output waypoints. Visibility deltas are reported strictly on cohort C3, the
$n=141$ subset on which both planners return a non-fallback trajectory; the
$107$ baseline-truncated scenarios are excluded because the 1-point
fallback yields no comparable trajectory and Town07 (cohort C4) is reported
separately because its visibility is dominated by environmental occlusion
rather than by the planner.

\subsection{Baseline configuration}
The priority-queue A* baseline shares the cost in Eq.~(\ref{eq:cost}) and
the feasibility constraints in Eq.~(\ref{eq:feasible}) with TA* but uses a
BinaryHeap-backed open set, no cross-time obstacle distance cache, no beam
pruning, the same $|\mathcal{N}| = 7$ pre-filtered neighbour set, and the
same expansion cap of $5{\times}10^{6}$. The 1-point fallback in
Eq.~(\ref{eq:fallback}) is the verbatim behaviour of that implementation
when the cap is reached.

\subsection{Town07 obstacle composition}
Table~\ref{tab:town07_obs} reports the full per-class obstacle composition
for cohort C4 referenced in \S\ref{subsec:town07-failure}. Three classes
(Fences, Poles, Vegetation) account for $73.9\%$ of records; among
classes that can plausibly occlude a $22\to 0$~m line-of-sight ray, the
fraction whose top altitude exceeds the $11$~m midpoint is
$77.8\%$ for Buildings, $38.5\%$ for Poles, and $31.2\%$ for Vegetation.

\begin{table}[htbp]
\caption{Town07 obstacle composition (cohort C4).}
\label{tab:town07_obs}
\centering
\small
\setlength{\tabcolsep}{4.5pt}
\begin{tabular}{lrrrr}
\toprule
\textbf{Class} & \textbf{Records} & \textbf{Share} & \textbf{zmax} & \textbf{$>$11\,m} \\
\midrule
Fences        & 9163  & 31.7\% & 1.3  & 0.0\% \\
Poles         & 7899  & 27.3\% & 12.2 & 38.5\% \\
Vegetation    & 4307  & 14.9\% & 6.5  & 31.2\% \\
Buildings     & 2109  & 7.3\%  & 20.6 & 77.8\% \\
Other$^{\dagger}$ & 5414 & 18.7\% & 1.3  & 0.0\% \\
\midrule
\textbf{All}  & 28892 & 100\%  & 6.5  & 20.9\% \\
\bottomrule
\end{tabular}\\[2pt]
{\footnotesize $^{\dagger}$Other = GuardRail + Terrain + TrafficSigns + Static + Bridge; ``zmax'' is the mean top altitude (m); ``$>$11\,m'' is the fraction whose top exceeds the line-of-sight midpoint of a 22\,m-to-0\,m ray.}
\end{table}

\bibliographystyle{IEEEtran}
\bibliography{references}

\end{document}